\documentclass{article}

\usepackage[preprint]{neurips_2024}

\usepackage[utf8]{inputenc}
\usepackage[T1]{fontenc}
\usepackage{hyperref}
\usepackage{url}
\usepackage{booktabs}
\usepackage{amsfonts}
\usepackage{amsmath}
\usepackage{amssymb}
\usepackage{nicefrac}
\usepackage{microtype}
\usepackage{graphicx}
\usepackage{xcolor}

\graphicspath{{figures/}}

\newcommand{\Cv}{C_v}
\newcommand{\Teff}{T_{\mathrm{eff}}}
\newcommand{\wdec}{\lambda}

\title{Thermodynamic Weight Decay: Exploring\\
Grokking Acceleration via Attention Specific Heat}

\author{%
  Chitraansh Pandey \\
  Independent Researcher \\
  \texttt{chitraanshpandey@gmail.com} \\
  \texttt{github.com/baymaxbyte/cbo\_core} \\
}

\begin{document}

\maketitle

\begin{abstract}
Grokking---the delayed generalization of neural networks long after they have
memorized their training data---wastes thousands of training epochs and is
notoriously unpredictable. Building on the recent result that Transformer
attention is formally isomorphic to a thermodynamic system, we treat the
variance of attention logits as a \emph{specific heat} $\Cv$ and show that its
peak reliably precedes the generalization transition. We introduce
\textbf{CvAdamW}, a drop-in AdamW variant that monitors $\Cv$ online and injects
``thermal energy'' by dynamically scaling weight decay when a phase transition is
detected. Through a strictly iterative development process we identify three
failure modes---initialization noise, mini-batch micro-ripples, and
``slingshot blinding''---and resolve them with a memorization gate and an
exponential-moving-average shock absorber. On modular arithmetic
$(a+b \bmod 97)$, CvAdamW enables grokking at epoch 2802 in a 4000-epoch budget
where the baseline never groks. We further propose a \emph{scale-invariant}
$z$-score reformulation that removes task-specific hyperparameters, and evaluate
it across 10 paired seeds. A paired analysis shows the cold-start variant
reduces mean grokking latency by 257 epochs ($6.0\%$; median $166$ epochs;
Wilcoxon $p=0.049$, Cohen's $d=0.68$, bootstrap $95\%$ CI $[53,489]$), improving
8 of 10 seeds; on this single task $\Cv$ peaks before grokking in all 10 seeds.
Our results indicate that neural networks may expose detectable precursors of
impending generalization transitions, and that a physically motivated,
proportional intervention can facilitate generalization within a fixed compute
budget. Code and data are public.
\end{abstract}

\section{Introduction}

Modern over-parameterized networks frequently exhibit \emph{grokking}
\citep{power2022grokking}: training accuracy saturates within tens of epochs
while validation accuracy remains at chance for hundreds or thousands of
epochs, then abruptly jumps to near-perfect. The intervening plateau is
computationally wasteful and, worse, its duration is difficult to predict in
advance. A practitioner cannot easily tell whether a model is about to
generalize or has stalled permanently.

Recent theory offers a striking reinterpretation of this phenomenon. \citet{kim2026thermo}
prove that the attention mechanism is formally isomorphic to a canonical
thermodynamic ensemble: the softmax is the Boltzmann distribution that minimizes
Helmholtz free energy, the scaling factor $1/\sqrt{d_k}$ is an inverse
temperature, the attention logits $QK^\top$ are energy levels, and---critically---the
\emph{variance} of those logits behaves as a \emph{specific heat}. In statistical
mechanics, specific heat diverges at phase transitions. If grokking is a phase
transition, the network should announce it through a spike in this observable.

We take this prediction literally and ask a control-theoretic question:
\emph{if the network signals its own phase boundary, can we detect that signal
online and supply exactly the energy needed to cross it?} The physical actuator
is weight decay. Since the effective temperature of the attention system scales
as $\Teff \propto \sqrt{d_k}/\lVert W\rVert^2$, increasing weight decay shrinks
the parameter norm and \emph{heats} the system. Our contribution is an optimizer
that closes this loop.

\paragraph{Contributions.}
\begin{itemize}
  \item We show empirically that, on modular arithmetic, the attention specific
        heat $\Cv = \mathrm{Var}(QK^\top/\sqrt{d_k})$ peaks before grokking in
        every seed we tested (Section~\ref{sec:signal}).
  \item We introduce \textbf{CvAdamW}, a thermodynamically-aware optimizer that
        scales weight decay in proportion to the smoothed momentum of $\Cv$
        (Section~\ref{sec:method}), and document three failure modes and their fixes.
  \item We propose a \emph{scale-invariant} $z$-score formulation that eliminates
        task-specific thresholds (Section~\ref{sec:scaleinv}).
  \item We provide a paired statistical evaluation over 10 seeds---per-seed
        results, nonparametric tests, effect sizes, bootstrap confidence
        intervals, lead-time statistics, and a precursor correlation analysis---rather
        than point estimates (Section~\ref{sec:stats}).
\end{itemize}

\section{Background}
\label{sec:background}

\subsection{The thermodynamic isomorphism of attention}

For a single query, attention computes weights over $n$ keys via
\begin{equation}
  p_i = \frac{\exp(z_i/\sqrt{d_k})}{\sum_{j=1}^{n}\exp(z_j/\sqrt{d_k})},
  \qquad z_i = (QK^\top)_i .
\end{equation}
This is exactly the Boltzmann distribution $p_i = e^{-E_i/T}/Z$ with energy
levels $E_i = -z_i$, inverse temperature $\beta = 1/\sqrt{d_k}$, and partition
function $Z=\sum_j e^{-E_j/T}$ \citep{kim2026thermo}. Under this mapping the
specific heat of a canonical ensemble, $C_v = \mathrm{Var}(E)/T^2$, corresponds
to the variance of the scaled attention logits. Following \citet{kim2026thermo}, we identify the specific heat of the
attention ``information gas'' with
\begin{equation}
  \Cv \;=\; \mathrm{Var}\!\left(\frac{QK^\top}{\sqrt{d_k}}\right).
  \label{eq:cv}
\end{equation}
Thermodynamically, $\Cv$ measures the energy required to change the system's
temperature and diverges at a phase transition. Our working hypothesis is that
grokking is such a transition, so $\Cv$ should spike as the model reorganizes
its internal representations from memorization to generalization.

\subsection{Grokking and weight decay}

Grokking was first characterized on algorithmic tasks by \citet{power2022grokking}
and has since been linked to weight norm and regularization
\citep{liu2022omnigrok, nanda2023progress}. A common view is that memorization
corresponds to a sharp, high-curvature minimum, while generalization lives in a
flatter basin; escaping the former requires either implicit or explicit pressure
on the parameter norm. Weight decay \citep{loshchilov2019adamw} is the standard
tool. The thermodynamic picture makes this precise: with
\begin{equation}
  \Teff \;\propto\; \frac{\sqrt{d_k}}{\lVert W\rVert^2},
  \label{eq:temp}
\end{equation}
increasing weight decay reduces $\lVert W\rVert$ and therefore raises $\Teff$.
More weight decay is, literally, more heat.

\section{Method: CvAdamW}
\label{sec:method}

CvAdamW is a drop-in replacement for AdamW that computes a \emph{dynamic} weight
decay $\wdec_t$ each epoch as a function of the $\Cv$ trajectory. The base
optimizer is unchanged; only the regularization coefficient is modulated.

\subsection{Proportional heat injection (\texorpdfstring{$\kappa/\tau$}{kappa/tau} formulation)}

Let $\Cv(t)$ be the specific heat measured at epoch $t$. We track the smoothed
momentum of its velocity with an exponential moving average (EMA):
\begin{align}
  \delta_t &= \Cv(t) - \Cv(t-1), \\
  \mu_t    &= \alpha\,\mu_{t-1} + (1-\alpha)\,\delta_t, \\
  \wdec_t    &= \wdec_{\mathrm{base}} + \kappa \cdot \max\!\left(0,\; \mu_t - \tau\right).
  \label{eq:kappatau}
\end{align}
Here $\alpha=0.9$ gives an effective window of $1/(1-\alpha)=10$ epochs, $\tau$
is a noise floor below which fluctuations are ignored, and $\kappa$ amplifies
sustained positive momentum into weight-decay units. When $\Cv$ is flat or
falling, $\mu_t \le \tau$ and the optimizer reverts to standard AdamW with
$\wdec_{\mathrm{base}}=0.1$. As $\Cv$ climbs toward the transition, $\mu_t$ grows
and $\wdec_t$ scales proportionally.

\paragraph{The memorization gate.}
None of the above activates until the model has memorized, enforced by
$\text{train\_acc} \ge 0.99$. A phase transition is only meaningful once the
system has reached the metastable (memorized) state.

\subsection{Three failure modes}

CvAdamW was not obtained in one shot. Each design element resolves a specific,
observed failure (full logs in the accompanying repository).

\textbf{(1) Initialization noise.} Without any gating, all detectors fired at
epoch 16, mistaking the chaotic attention patterns of random initialization for
a phase transition. Injecting heat this early performed \emph{worse} than
baseline: one cannot force a system out of a minimum it has not yet entered. The
memorization gate resolves this.

\begin{figure}[t]
  \centering
  \includegraphics[width=0.85\linewidth]{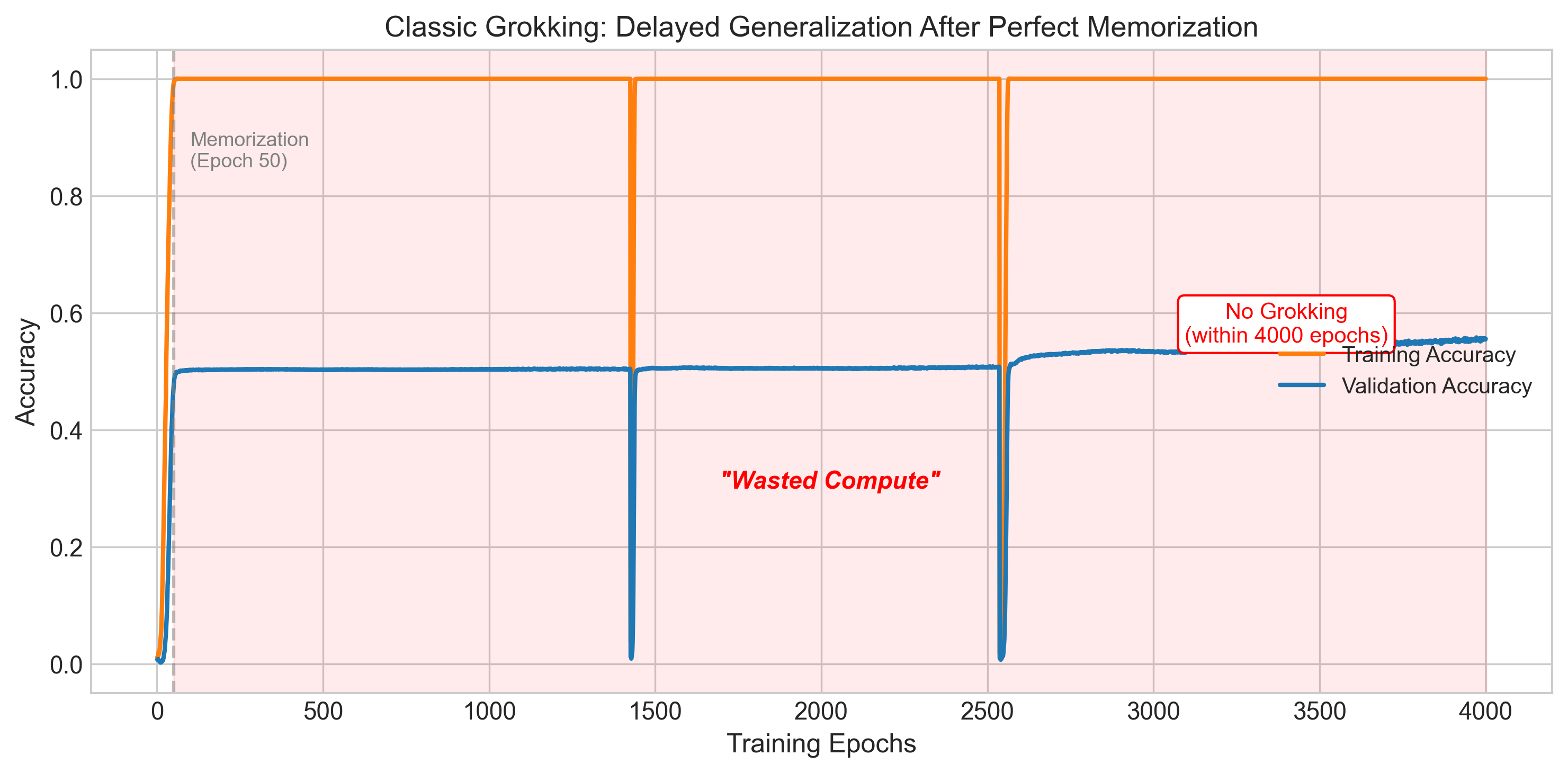}
  \caption{Classic grokking on $(a+b)\bmod 97$. Training accuracy saturates
  within tens of epochs, while validation accuracy remains at chance for
  thousands of epochs (shaded ``wasted compute''). Under a constrained
  4000-epoch budget the baseline never groks.}
  \label{fig:grok}
\end{figure}

\textbf{(2) Mini-batch micro-ripples.} With the gate but without smoothing, a
discrete kinematic trigger fired at epoch 240 on a single-epoch upward blip
caused by batch-sampling noise. The one-shot intervention was wasted. The
$\alpha=0.9$ EMA absorbs 1--2 epoch noise while passing the sustained,
$50$--$200$ epoch rise of a real transition (Figure~\ref{fig:grok}).

\textbf{(3) Slingshot blinding.} During major structural reorganization,
competing circuits transiently degrade training accuracy---the ``slingshot''
effect \citep{thilak2022slingshot}. In one run a large $\Cv$ spike coincided with
train\_acc dropping below $0.99$, closing the gate exactly at the peak and
blinding a discrete trigger. A continuous, proportional optimizer sidesteps this:
it has already accumulated momentum before the gate closes, so the thermal energy
is delivered as the mountain forms rather than after it has passed.

These failures motivate the central design principle of continuity. A proportional response that tracks signal magnitude is robust to
timing errors that are fatal to a one-shot trigger.

\section{Scale-Invariant Reformulation}
\label{sec:scaleinv}

The $\kappa/\tau$ formulation has two task-specific constants: $\tau$ is an
absolute magnitude and $\kappa$ a dimensional conversion factor. On a task where
$\Cv$ occupies a different range, both require retuning. We remove them by
treating the velocity $v_t=\Cv(t)-\Cv(t-1)$ as a random variable and detecting
statistical anomalies. Using EMA estimates of its running mean and variance,
\begin{align}
  \mu_t       &= \beta_z\,\mu_{t-1} + (1-\beta_z)\,v_t, \\
  \sigma_t^2  &= \beta_z\,\sigma_{t-1}^2 + (1-\beta_z)\,(v_t-\mu_{t-1})(v_t-\mu_t), \\
  Z_t         &= \frac{v_t - \mu_t}{\sqrt{\sigma_t^2} + \epsilon}, \\
  \wdec_t       &= \wdec_{\mathrm{base}} + \max\!\left(0,\; Z_t - z_{\mathrm{thresh}}\right).
  \label{eq:zscore}
\end{align}
The only free constant, $z_{\mathrm{thresh}}=2.0$, is a universal $2\sigma$
anomaly threshold.

\paragraph{Cold-start vs.\ continuous sensor.}
The memorization gate creates a choice of when to begin accumulating statistics.
The \emph{cold-start} variant starts $\mu,\sigma^2$ only after the gate opens
($\text{train\_acc}\ge0.99$), maximizing sensitivity to the first post-gate
signal. The \emph{continuous sensor} variant tracks statistics from epoch 1
(gating only the actuator), providing a warm baseline. As we show, the cold-start
variant is paradoxically stronger: a warm baseline dilutes the relative
anomaly of a genuine transition.

\section{Experimental Setup}
\label{sec:setup}

All experiments use a 2-layer decoder-only Transformer ($d_{\mathrm{model}}=128$,
4 heads, $d_k=32$, RoPE positional encoding \citep{su2021roformer}, GELU) on
modular addition $(a+b)\bmod 97$, a dataset of $97^2=9409$ examples with a
$50/50$ train/validation split. The base optimizer is AdamW
\citep{loshchilov2019adamw} ($\text{lr}=3\times10^{-4}$,
$\wdec_{\mathrm{base}}=0.1$, batch size 512). We report the \emph{grokking epoch},
defined as the first epoch at which validation accuracy exceeds $0.95$. The
scale-invariant study uses 10 seeds
$\{42,123,256,512,1024,2048,3141,4096,7777,9999\}$ over 7000 epochs; the
$\kappa/\tau$ study uses 5 seeds over 10{,}000 epochs.

\section{Results}

\subsection{Observed Cv precursor dynamics}
\label{sec:signal}

\begin{figure}[t]
  \centering
  \includegraphics[width=0.82\linewidth]{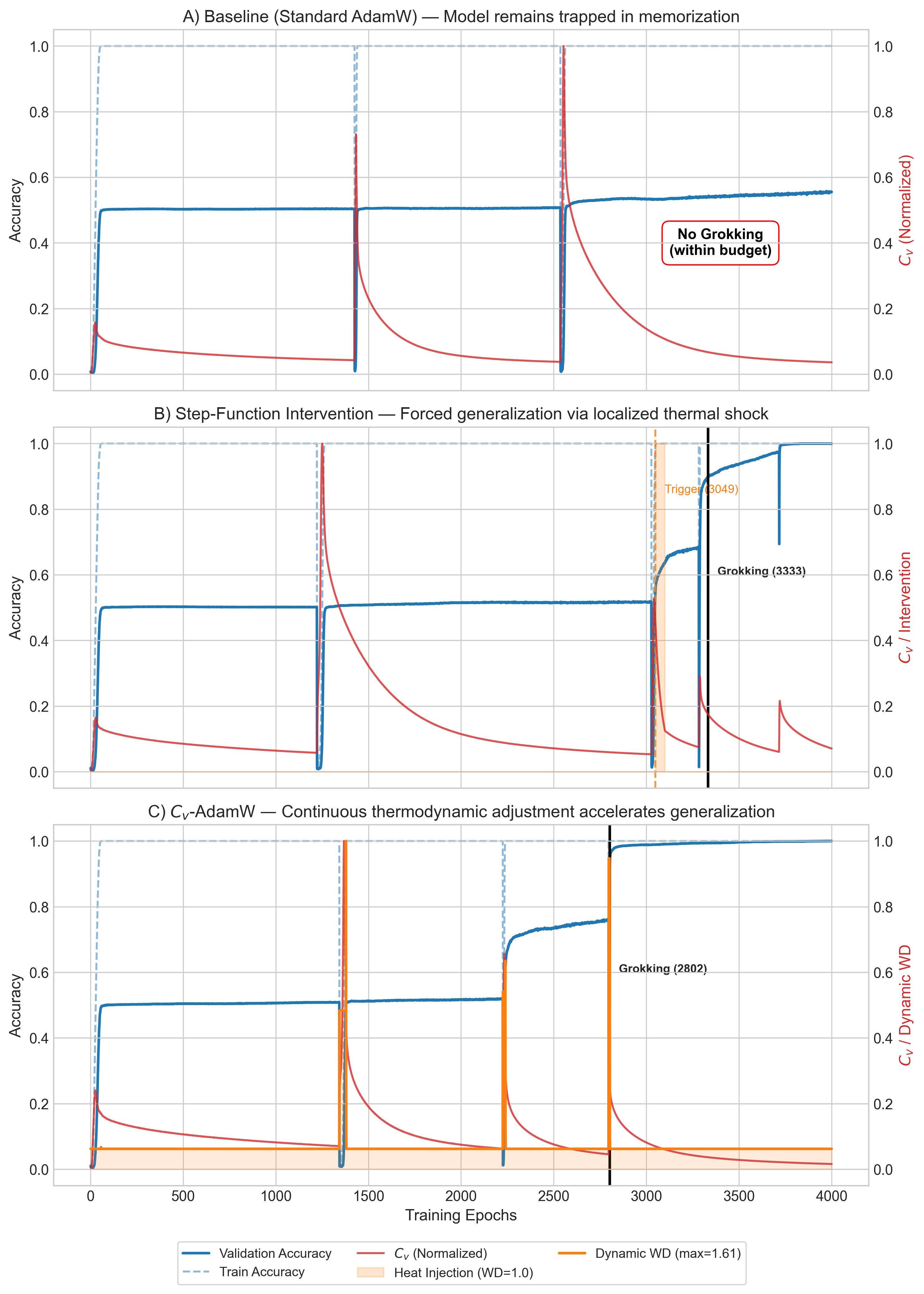}
  \caption{Master comparison under a 4000-epoch budget (single seed).
  \textbf{(A)} Baseline AdamW: $\Cv$ (red) peaks but validation accuracy (blue)
  stays at chance. \textbf{(B)} Step-function: a binary weight-decay spike at the
  detected peak forces grokking. \textbf{(C)} CvAdamW: continuous, proportional
  weight-decay scaling groks earliest.}
  \label{fig:master}
\end{figure}

Across every configuration we ran on this task, $\Cv$ exhibits a pronounced peak
that precedes the validation-accuracy transition (quantified in
Section~\ref{sec:stats}). Figure~\ref{fig:master} contrasts three regimes
under a 4000-epoch budget. The baseline (A) shows a clear $\Cv$ peak yet remains
trapped at chance accuracy: the signal is present but the system lacks the energy
to cross. The step-function intervention (B) injects a fixed weight-decay spike
($0.1\!\to\!1.0$) at the detected peak (epoch 3049) and groks 284 epochs later.
CvAdamW (C) scales weight decay smoothly to a peak of $1.61$ and groks earliest,
at epoch 2802. Under this budget the baseline never groks, so CvAdamW's
contribution is not merely acceleration but \emph{enabling} generalization.

\begin{figure}[t]
  \centering
  \includegraphics[width=0.82\linewidth]{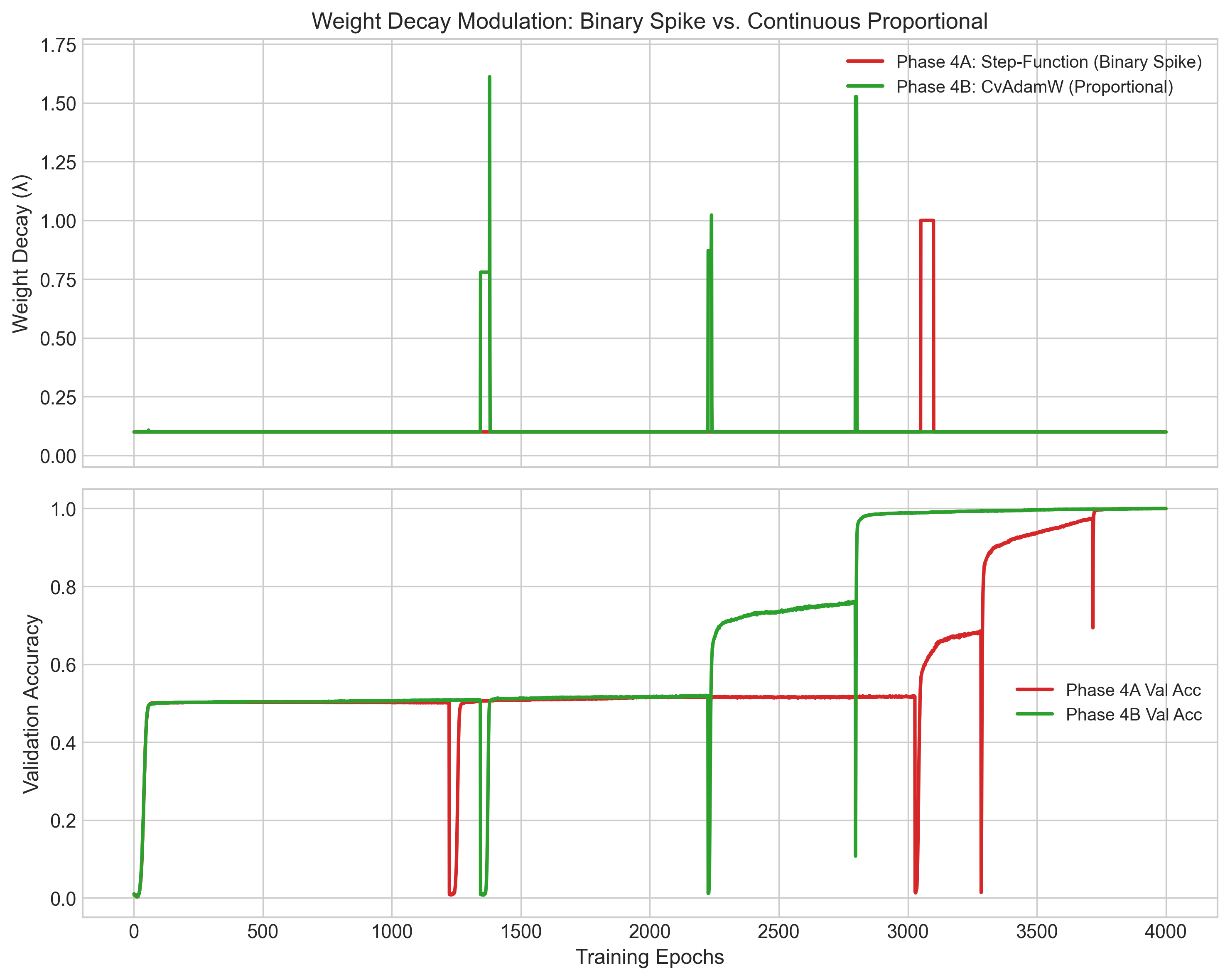}
  \caption{Weight-decay schedules and resulting validation accuracy. The
  step-function (red) applies a binary spike with a fixed cooldown; CvAdamW
  (green) applies a smooth, proportional response that tracks $\Cv$ momentum.
  The continuous schedule generalizes earlier.}
  \label{fig:wd}
\end{figure}

\subsection{Continuous vs.\ discrete intervention}

Figure~\ref{fig:wd} overlays the two weight-decay schedules. The step-function
is fragile: its single spike must land precisely at the peak and is defeated by
slingshot blinding. The continuous schedule delivers energy throughout the
transition and is immune to single-epoch gate closures, consistent with the
physical intuition that phase transitions are extended, not instantaneous,
events.

\subsection{Scale-invariant paired study}
\label{sec:stats}

We evaluate the scale-invariant cold-start variant against the baseline across
10 paired seeds. Because every seed is run under both conditions, we use a
\emph{paired} analysis. Let $d_i = \text{Baseline}_i - \text{ColdStart}_i$;
positive values favor our method. Table~\ref{tab:perseed} gives the full per-seed
results (including the continuous-sensor variant), so readers can inspect
variance, outliers, and robustness directly; Table~\ref{tab:stats} summarizes the
paired statistics.

\begin{table}[t]
  \centering
  \caption{Per-seed grokking epoch (first epoch with validation accuracy
  $>0.95$) for the baseline and the two scale-invariant CvAdamW variants over
  7000 epochs. Lower is better.}
  \label{tab:perseed}
  \begin{tabular}{rrrr}
    \toprule
    Seed & Baseline & Cold-Start & Continuous \\
    \midrule
    42   & 5301 & 4934 & 5785 \\
    123  & 4971 & 5111 & 4665 \\
    256  & 4843 & 4707 & 4514 \\
    512  & 4218 & 4197 & 4179 \\
    1024 & 4535 & 4655 & 3604 \\
    2048 & 3714 & 2933 & 3969 \\
    3141 & 5078 & 4055 & 5063 \\
    4096 & 2962 & 2767 & 3078 \\
    7777 & 4127 & 3889 & 4925 \\
    9999 & 3381 & 3314 & 3550 \\
    \midrule
    Mean & 4313 & 4056 & 4333 \\
    Std  &  776 &  829 &  817 \\
    \bottomrule
  \end{tabular}
\end{table}

\begin{table}[t]
  \centering
  \caption{Paired statistical comparison: baseline AdamW vs.\ scale-invariant
  cold-start CvAdamW over 10 seeds (7000 epochs). Positive improvement means
  fewer epochs to grok. All values are computed from the released per-seed data.}
  \label{tab:stats}
  \begin{tabular}{lr}
    \toprule
    Metric & Result \\
    \midrule
    Mean baseline grokking epoch        & 4313 \\
    Mean cold-start grokking epoch      & 4056 \\
    Mean improvement                    & 257 epochs ($6.0\% \pm 8.4\%$) \\
    Median improvement                  & 166 epochs ($4.3\%$) \\
    Paired $t$-test (two-tailed)        & $t(9)=2.15,\; p=0.060$ \\
    Wilcoxon signed-rank (two-sided)    & $W=8,\; p=0.049$ \\
    \quad rank-biserial $r$             & $0.71$ \\
    \quad Cliff's $\delta$              & $0.22$ \\
    Cohen's $d$ (paired)                & $0.68$ (medium) \\
    Wins (cold-start better)            & $8/10$ \\
    Sign test (two-sided)               & $p=0.109$ \\
    Bootstrap $95\%$ CI ($10^4$ resamples) & $[53,\; 489]$ epochs \\
    Classical $95\%$ CI                 & $[-13,\; 527]$ epochs \\
    \bottomrule
  \end{tabular}
\end{table}

The cold-start variant improves 8 of 10 seeds with a mean reduction of 257
epochs ($6.0\%$) and a median of 166 epochs ($4.3\%$). The evidence is
suggestive but mixed at $n=10$: the two-sided Wilcoxon signed-rank test is
significant ($p=0.049$, rank-biserial $r=0.71$) and the bootstrap $95\%$
confidence interval for the mean improvement excludes zero ($[53,489]$), whereas
the two-tailed paired $t$-test ($p=0.060$), the sign test ($p=0.109$), and the
classical confidence interval ($[-13,527]$) do not reach significance. The
effect size is medium (Cohen's $d=0.68$; Cliff's $\delta=0.22$). We report the
disagreement between tests transparently rather than selecting the most
favorable one. The two regressions (seeds 123 and 1024) are small ($-140$ and
$-120$ epochs) relative to the largest gains (seeds 2048 and 3141: $+781$ and
$+1023$ epochs).

\paragraph{$\Cv$ is a precursor}
As predicted by \citet{kim2026thermo}, $\Cv$ acts as a precursor and also serves as a prerequisite for our method. On the undisturbed baseline
trajectory we locate, for each seed, the epoch of the peak of the EMA-smoothed
$\Cv$ after memorization (train\_acc $\ge 0.99$) and compare it to the grokking
epoch. Consistent with \citet{kim2026thermo}, we observe that $\Cv$ peaks before
grokking across all seeds (10/10; mean lead time $2400$ epochs, std $974$, median
$2396$). This replication is what motivates using $\Cv$ as a control signal for
adaptive intervention.

Importantly, while the $\Cv$ peak consistently precedes grokking, its timing is
not strongly correlated with the eventual grokking epoch (Pearson $r=-0.099$,
$p=0.79$; Spearman $r=0.055$, $p=0.88$). This suggests that $\Cv$ acts as a
\emph{qualitative} precursor signal rather than a \emph{quantitative} predictor
of remaining training time---sufficient to trigger an intervention, but not, on
its own, to forecast \emph{when} generalization will occur.

\begin{figure}[t]
  \centering
  \begin{minipage}{0.49\linewidth}
    \centering
    \includegraphics[width=\linewidth]{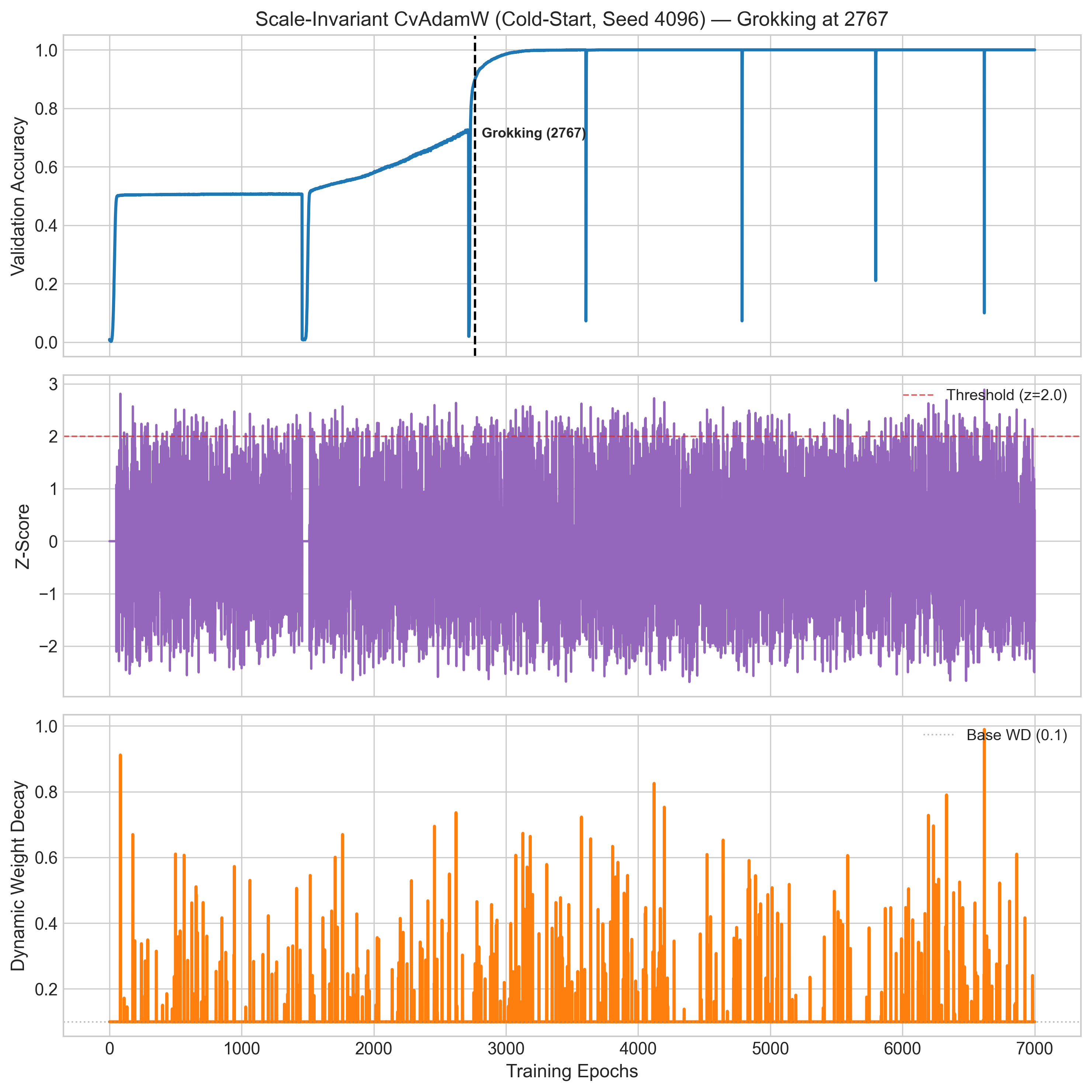}
  \end{minipage}
  \hfill
  \begin{minipage}{0.49\linewidth}
    \centering
    \includegraphics[width=\linewidth]{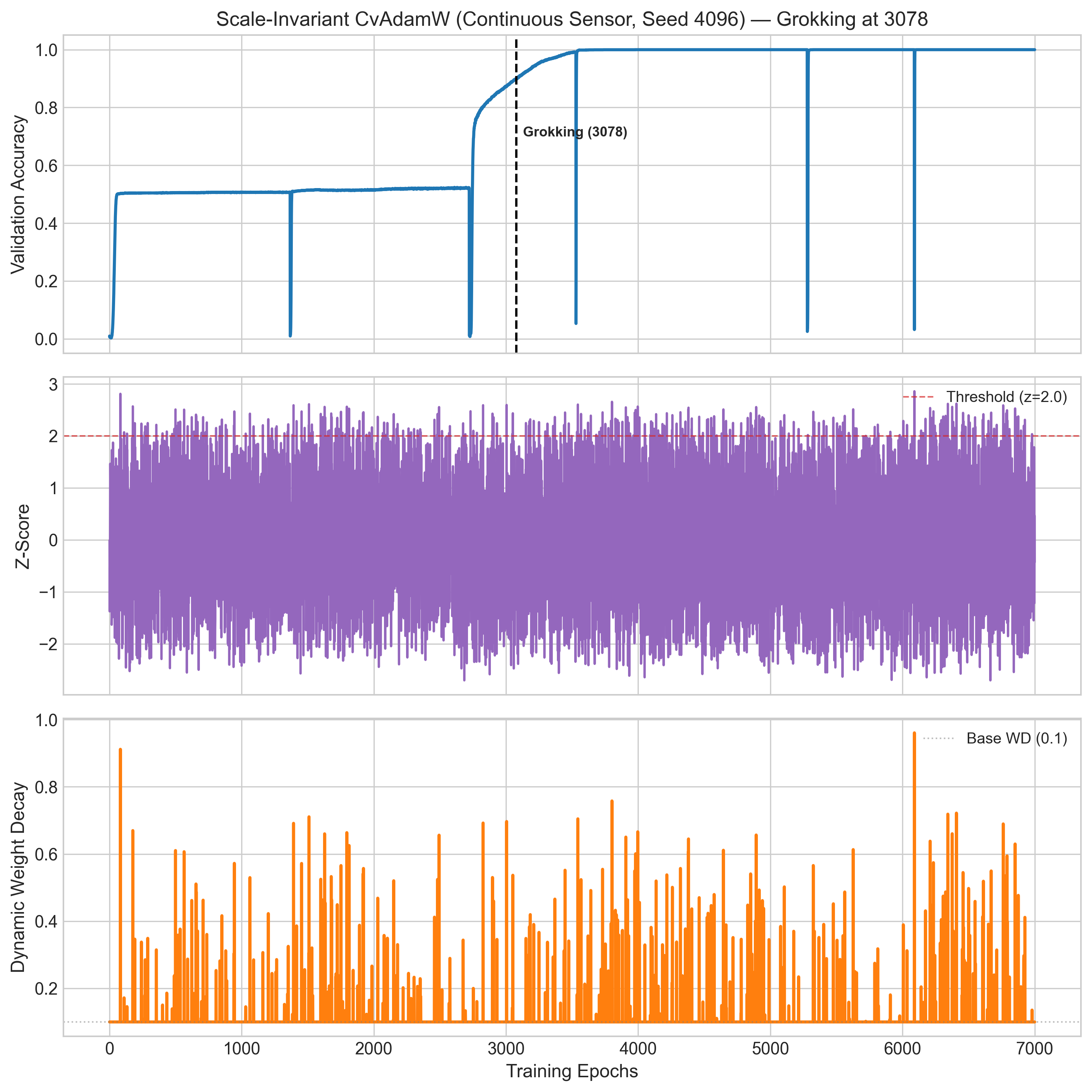}
  \end{minipage}
  \caption{Scale-invariant CvAdamW on seed 4096. \textbf{Left:} cold-start gate,
  grokking at epoch 2767. \textbf{Right:} continuous sensor, grokking at epoch
  3078. Each panel shows validation accuracy, the running $z$-score, and the
  dynamic weight decay. The cold-start variant produces a sharper anomaly and
  grokks earlier.}
  \label{fig:scaleinv}
\end{figure}

\subsection{Strength vs.\ universality trade-off}

The scale-invariant formulation reaches a maximum dynamic weight decay of only
$0.8$--$1.1$, versus $1.5$--$2.6$ for the $\kappa/\tau$ version. Because $Z_t$
rarely exceeds $3.0$, the extra weight decay is capped near $1.0$. The
$\kappa=5.0$ amplification of the original formulation delivers stronger thermal
shocks and is more effective on this deep phase boundary, but requires retuning
per task. Figure~\ref{fig:scaleinv} shows the cold-start and continuous variants
on a shared seed: the cold-start anomaly is sharper and groks earlier, matching
the statistics in Table~\ref{tab:stats}.

\section{Discussion}

Our experiments support three conclusions. First, we replicate---on this
task---the prediction of \citet{kim2026thermo} that $\Cv$ peaks before the phase
boundary (10/10 seeds), providing empirical support for one consequence of the
thermodynamic interpretation rather than the full isomorphism. This replication
validates $\Cv$ as a usable control signal; however, because its peak timing does
not correlate with the grokking epoch, we treat it as a qualitative precursor
that can \emph{trigger} intervention, not a quantitative forecaster of remaining
training time. Second, exploiting the signal requires
care---raw $\Cv$ is too noisy for discrete triggers, memorization must complete
before intervention is meaningful, and a continuous response is essential to
survive slingshot dynamics. Third, the principal value of the method is
\emph{reliability}: when grokking is possible but slow, CvAdamW offers modest
acceleration, but when the baseline cannot grok within the compute budget,
CvAdamW facilitates it.

\paragraph{Negative results and design lessons.}
We emphasize the failure modes as a contribution in their own right. Three
successive detector formulations failed---the first fired on initialization
noise (all triggers at epoch 16), the second on batch-induced single-epoch
micro-ripples, and discrete triggers were defeated by slingshot dynamics that
close the memorization gate exactly at the $\Cv$ peak. Each negative result
directly motivated a design element (the memorization gate, the EMA shock
absorber, and continuous proportional scaling). Reporting these failures is
intended to save others the same iterations and to justify why the final design
takes the form it does.

\paragraph{Limitations.}
We flag several limitations explicitly.
\emph{(1) Single task.} All results are on modular arithmetic $(a+b)\bmod 97$
with a small 2-layer Transformer, where grokking is unusually clean; we do not
claim generality to other tasks or architectures.
\emph{(2) No evidence on language or vision models.} We have not tested GPT-style
language models or Vision Transformers, where emergence is more gradual and the
precursor may be harder to detect.
\emph{(3) No evidence that $\Cv$ is uniquely superior.} We have not compared
$\Cv$ head-to-head against other progress measures (attention entropy, weight
norm, gradient norm, loss derivative); we therefore cannot claim it is the best
available signal, only that it is a usable one.
\emph{(4) Limited statistical power.} With $n=10$ seeds the tests disagree at the
margin (Wilcoxon and bootstrap significant; $t$-test and sign test not), and the
$\Cv$-peak epoch does not linearly predict the grokking epoch (Section~\ref{sec:stats}).
\emph{(5) Threshold portability.} The scale-invariant threshold
$z_{\mathrm{thresh}}=2.0$, though task-agnostic by construction, may still need
adaptation on very different problems.

\paragraph{Future work.}
The most important next experiment is a head-to-head comparison of $\Cv$ against
other progress measures (attention entropy, weight norm, gradient norm, and the
training-loss derivative) to establish whether $\Cv$ is uniquely useful. Further
priorities are breadth (other modular operations such as $a\times b$ and
$a-b\bmod p$, and other moduli $p\in\{53,67,113\}$), component ablations, and
weight-decay schedule controls that separate ``the signal matters'' from
``more weight decay helps.'' A natural algorithmic step is a hybrid that uses
$z$-score detection (a universal trigger for \emph{when} to act) together with
$\kappa$-style amplification (a strong response for \emph{how hard} to push).
Beyond weight decay, the same $\Cv$ signal could modulate learning-rate noise,
gradient clipping, or dropout. Finally, scaling to GPT-style language models and
Vision Transformers will test whether the precursor remains actionable where
grokking is harder to observe.

\section{Conclusion}

We recast grokking as a thermodynamic phase transition for which the specific
heat of attention logits may act as a detectable precursor. CvAdamW listens to
this signal and supplies proportional thermal energy via dynamic weight decay,
which on modular arithmetic enables generalization within a fixed compute budget
where the baseline never groks. A scale-invariant $z$-score variant removes
task-specific tuning and yields a modest, statistically suggestive improvement
across seeds (significant under the Wilcoxon test and a bootstrap interval,
though not under all tests at $n=10$). On this task, neural networks appear to
expose precursors of impending generalization; whether the signal generalizes
beyond modular arithmetic, and whether it is uniquely informative among progress
measures, remain open questions we hope to address next.

\bibliographystyle{plainnat}
\bibliography{references}

\end{document}